\documentclass[11pt,a4paper]{article}
\usepackage[hyperref]{emnlp-ijcnlp-2019}
\usepackage{times}
\usepackage{latexsym}
\usepackage{subcaption}
\usepackage{graphicx}
\usepackage{amsmath}
\usepackage{mathtools}
\usepackage{url}
\usepackage{booktabs} 

\aclfinalcopy

\title{Feature-Dependent Confusion Matrices for Low-Resource NER Labeling with Noisy Labels}

\author{Lukas Lange$^{1,2}$\\
  \And
  Michael A. Hedderich$^2$\\
  $^1$ Bosch Center for Artificial Intelligence, Renningen, Germany \\
  $^2$ Spoken Language Systems (LSV), Saarland University,\\ Saarland Informatics Campus, Saarbr\"{u}cken, Germany\\
  {\tt \{llange,mhedderich,dietrich.klakow\}@lsv.uni-saarland.de} \\\And
  Dietrich Klakow$^2$\\
\\}

\date{}

\begin{document}
\maketitle

\begin{abstract}
In low-resource settings, the performance of supervised labeling models can be improved with automatically annotated or distantly supervised data, which is cheap to create but often noisy.
Previous works have shown that significant improvements can be reached by injecting information about the confusion between clean and noisy labels in this additional training data into the classifier training. 
However, for noise estimation, these approaches either do not take the input features (in our case word embeddings) into account, or they need to learn the noise modeling from scratch which can be difficult in a low-resource setting. 
We propose to cluster the training data using the input features and then compute different confusion matrices for each cluster. To the best of our knowledge, our approach is the first to leverage feature-dependent noise modeling with pre-initialized confusion matrices.
We evaluate on low-resource named entity recognition settings in several languages, showing that our methods improve upon other confusion-matrix based methods by up to 9\%. 
\end{abstract}

\section{Introduction}\label{sec:intro}
Most languages, even with millions of speakers, have not been the center for natural language processing and are counted as low-resource for tasks like named entity recognition (NER). Similarly, even for high-resource languages, there exists only few labeled data for most entity types beyond person, location and organization. Distantly- or weakly-supervised approaches
have been proposed to solve this issue, e.g., by using lists of entities for labeling raw text \cite{ner/Ratinov2009, ner/Dembowski2017}. This allows obtaining large amounts of training data quickly and cheaply. Unfortunately, these labels often contain errors and learning with this noisily-labeled data is difficult and can even reduce overall performance (see, e.g. \citet{noise/Fang16}).

A variety of ideas have been proposed to overcome the issues of noisy training data. One popular approach is to estimate the relation between noisy and clean, gold-standard labels and use this noise model to improve the training procedure. However, most of these approaches only assume a dependency between the labels and do not take the features into account when modeling the label noise. This may disregard important information. 
The global confusion matrix~\cite{noise/Hedderich18} is a simple model which assumes that the errors in the noisy labels just depend on the clean labels. 

Our contributions are as follows: We propose to cluster the input words with the help of additional, unlabeled data. Based on this partition of the feature space, we obtain different confusion matrices that describe the relationship between clean and noisy labels. We evaluate our newly proposed models and related baselines in several low-resource settings across different languages with real, distantly supervised data with non-synthetic noise.
The advanced modeling of the noisy labels substantially improves the performance up to 36\% over methods without noise-handling and up to 9\% over all other noise-handling baselines. 

\section{Related Work}
A popular approach is modeling the relationship between noisy and clean labels, i.e., estimating $p(\hat{y}|y)$ where $y$ is the clean and $\hat{y}$ the noisy label. For example, this can be represented as a noise or confusion matrix between the clean and the noisy labels, as explained in Section~\ref{sec:global}. Having its roots in statistics \cite{noise/Dawid79}, this or similar ideas have been recently studied in NLP \cite{noise/Fang16, noise/Hedderich18, noise/Paul19}, image classification \cite{noise/Mnih2012, noise/Sukhbaatar14, noise/Dgani2018} and general machine learning settings \cite{noise/Bekker16, noise/Patrini2017, noise/Hendrycks2018}. All of these methods, however, do not take the features into account 
that are used to represent the instances during classification.
In \cite{noise/Xiao2015} only the noise type depends on $x$ but not the actual noise model. \citet{noise/Goldberger16} and \citet{noise/Luo17} use the learned feature representation $h$ to model $p(\hat{y}|y,h(x))$ for image classification and relation extraction respectively. In the work of \citet{noise/Veit17}, $p(y|\hat{y},h(x))$ is estimated to clean the labels for an image classification task. The survey by \citet{noise/Frenay2014} gives a detailed overview about other techniques for learning in the presence of noisy labels. 

Specific to learning noisy sequence labels in NLP, \citet{noise/Fang16} used a combination of clean and noisy data for low-resource POS tagging. \citet{noise/Yang18} suggested partial annotation learning to lessen the effects of incomplete annotations and reinforcement learning for filtering incorrect labels for Chinese NER. \citet{noise/Hedderich18} used a confusion matrix and proposed to leverage pairs of clean and noisy labels for its initialization, evaluating on English NER. For English NER and Chunking, \citet{noise/Paul19} also used a confusion matrix but learned it with an EM approach and combined it with multi-task learning. Recently, \citet{noise/Rahimi19} studied input from different, unreliable sources and how to combine them for NER prediction.

\section{Global Noise Model}
\label{sec:global}
We assume a low-resource setting with a small set of gold standard annotated data $C$ consisting of instances with features $x$ and corresponding, clean labels $y$. Additionally, a large set of noisy instances $(x,\hat{y}) \in N$ is available. This can be obtained e.g. from weak or distant supervision. In a multi-class classification setting, we can learn the probability of a label $y$ having a specific class given the feature $x$ as

\begin{align}
    p(y=i|x) = \frac{exp(u_i^Th(x))}{\sum_{l=1}^k exp(u_l^Th(x))}
\end{align}

where $k$ is the number of classes, $h$ is a learned, non-linear function (in our case a neural network) and $u$ is the softmax weights. This is our base model trained on $C$. Due to the errors in the labels, the clean and noisy labels have different distributions. Therefore, learning on $C$ and $N$ jointly can be detrimental for the performance of predicting unseen, clean instances. Nevertheless, the noisy-labeled data is still related to $C$ and can contain useful information that we want to successfully leverage. 
We transform the predicted (clean) distribution of the base model to the noisy label distribution

\begin{align}
p(\hat{y}=j|x) = \sum^k_{i=1} p(\hat{y}=j| y=i)p(y=i|x). \label{eq:noise-prediction}
\end{align}

The relationship is modeled using a confusion matrix (also called noise or transformation matrix or noise layer) with learned weights $b_{ij}$:

\begin{align}
    p(\hat{y}=j|y=i) = \frac{\exp(b_{ij}) }{\sum_{l=1}^k \exp(b_{il})} \label{eq:noise-clean-softmax}
\end{align}

\begin{figure}
    \centering
    \includegraphics[width=0.4\textwidth]{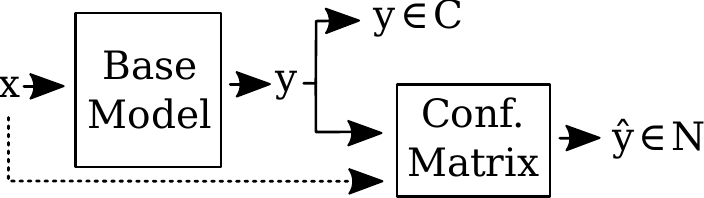}
    \caption{Visualization of the noisy labels, confusion matrix architecture. The dotted line shows the proposed new dependency.}
    \label{fig:noise_architecture}
\end{figure}

The overall architecture is visualized in Figure \ref{fig:noise_architecture}. An important question is how to initialize this noise layer. As proposed by \citet{noise/Hedderich18}, we apply the same distant supervision technique used to obtain $N$ from unlabeled data on the already labeled instances in $C$. We thus obtain pairs of clean $y$ and corresponding noisy labels $\hat{y}$ for the same instances and the weights of the noise layer can be initialized as

\begin{align}
    b_{ij} = \log(\frac{\sum_{t=1}^{|C|} 1_{\{y_t=i\}} 1_{\{\hat{y}_t=j\}}}{\sum_{t=1}^{|C|} 1_{\{y_t=i\}}}) . \label{eq:global-init}
\end{align}

Following the naming by \cite{noise/Luo17}, we call this the global noise model.

\section{Feature Dependent Noise Model}
\label{sec:feature-model}

The global confusion matrix is a simple model which assumes that the errors in the noisy labels depend on the clean labels. An approach that also takes the corresponding features $x$ into account can model more complex relations. \citet{noise/Veit17} and \citet{noise/Luo17} use multiple layers of a neural network to model these relationships. However, in low resource settings with only small amounts of clean, supervised data, these more complex models can be difficult to learn. In contrast to that, larger amounts of unlabeled text are usually available even in low-resource settings. Therefore, we propose to use unsupervised clustering techniques to partition the feature space of the input words (and the corresponding instances) before estimating the noise matrices. To create the clusters, we use either Brown clustering~\citep{cluster/Brown92} on the input words or $k$-means clustering~\citep{cluster/Lloyd82} on the pretrained word embeddings after applying PCA~\cite{pca/Pearson1901}.

In sequence labeling tasks, the features $x$ of an instance usually consist of the input word $\iota(x)$ and its context. Given a clustering $\Pi$ over the input words $\{\iota(x) \mid (x,y) \in C \cup N \}$ consisting of clusters $\Pi_1, ..., \Pi_p$, we can group all clean and noisy instances into groups 

\begin{align}
    G_q = \{ (x,y) \in C \cup N \mid \iota(x) \in \Pi_q \}
\end{align}

For each group, we construct an independent confusion matrix using Formulas \ref{eq:noise-clean-softmax} and \ref{eq:global-init}. The prediction of the noisy label $\hat{y}$ (Formula \ref{eq:noise-prediction}) then becomes

\begin{align}
    p(\hat{y}=j|x) = \sum^k_{i=1} p(\hat{y}=j| y=i, G)p(y=i|x)
\end{align}

Since the clustering is performed on unsupervised data, in low-resource settings, the size of an actual group of instances $G_q$ can be very small. If the number of members in a group is insufficient, the estimation of reliable noise matrices is difficult. This issue can be avoided by only using the largest groups and creating a separate group for all other instances. To make use of all the clusters, we alternatively propose to interpolate between the global and the group confusion matrix:

\begin{multline}
  p_{\text{int}}(\hat{y}=j| y=i, G) = \\
  (1-\lambda) \cdot p(\hat{y}=j| y=i, G) + \lambda \cdot p(\hat{y}=j| y=i) \quad
\end{multline}

The interpolation hyperparameter $\lambda$ (with $0 \leq \lambda \leq 1$) regulates the influence from the global matrix on the interpolated matrix. The selection of the largest groups and the interpolation can also be combined. 

\begin{table*}[ht!]
    \centering
    \footnotesize
    \begin{tabular}{l|ccccc} \toprule
         & De & En & Es & Et & Nl \\ \midrule
         
        Base
        & 21.4 $\pm$ 1.0 & 35.9 $\pm$ 4.6 & 39.1 $\pm$ 1.6 & 36.7 $\pm$ 1.8 & 15.5 $\pm$ 3.0 \\
        
        Base+Noise 
        & 26.2 $\pm$ 0.6 & 50.5 $\pm$ 1.4 & 50.2 $\pm$ 1.0 & 51.5 $\pm$ 0.7 & 29.5 $\pm$ 2.7 \\ \midrule
        
        Cleaning (Veit et al. 2017)
        & 16.1 $\pm$ 4.3 & 52.3 $\pm$ 2.3 & 48.7 $\pm$ 2.3 & 53.8 $\pm$ 0.4 & 24.4 $\pm$ 5.5 \\
        
        Dynamic-CM (Luo et al. 2017)
        & 32.6 $\pm$ 0.9 & 53.7 $\pm$ 1.8 & 57.6 $\pm$ 0.8 & 52.3 $\pm$ 0.8 & 36.7 $\pm$ 2.9 \\
        
        Global-ID-CM (H. and K. 2018)
        & 27.1 $\pm$ 0.7 & 51.0 $\pm$ 1.1 & 50.9 $\pm$ 0.7 & 51.4 $\pm$ 0.6 & 29.9 $\pm$ 2.6 \\
        
        Global-CM (H. and K. 2018)
        & 34.1 $\pm$ 1.4 & 52.0 $\pm$ 1.6 & 52.8 $\pm$ 0.6 & 52.3 $\pm$ 0.6 & 33.3 $\pm$ 2.0 \\ \midrule
        
        Brown-CM-Freq
        & 32.7 $\pm$ 0.7 & 51.3 $\pm$ 1.3 & 54.8 $\pm$ 1.0 & 53.4 $\pm$ 0.8 & 38.1 $\pm$ 1.7 \\
        
        K-Means-CM-Freq
        & 29.7 $\pm$ 2.3 & 54.1 $\pm$ 2.9 & 52.3 $\pm$ 1.2 & 54.9 $\pm$ 0.8 & 39.8 $\pm$ 1.8 \\
        
        Brown-CM-IP
        & 29.6 $\pm$ 1.1 & 55.5 $\pm$ 3.7 & 55.6 $\pm$ 1.0 & 52.6 $\pm$ 0.9 & 37.3 $\pm$ 1.5 \\
        
        K-Means-CM-IP
        & 33.4 $\pm$ 1.1 & 53.0 $\pm$ 4.0 & 56.3 $\pm$ 2.1 & 53.3 $\pm$ 0.5 & 36.0 $\pm$ 1.9 \\ \midrule
        
        Brown-CM-Freq-IP
        & \textbf{34.3 $\pm$ 1.4} & 51.4 $\pm$ 2.3 & \textbf{57.7 $\pm$ 2.4} & 53.1 $\pm$ 0.9 & \textbf{40.0 $\pm$ 1.3} \\
        
        K-Means-CM-Freq-IP
        & 33.1 $\pm$ 2.1 & \textbf{57.6 $\pm$ 1.5} & 57.2 $\pm$ 1.3 & \textbf{55.2 $\pm$ 0.3} & 39.7 $\pm$ 1.0 \\ \bottomrule
        
    \end{tabular}
    \caption{Results of the evaluation in low-resource settings with 1\% of the original labeled training data averaged over six runs. We report the F1 scores (higher is better) on the complete test set, as well as the standard error.}
    \label{tab:results}
\end{table*}

\begin{figure}
  \centering
    \begin{subfigure}[t]{0.19\textwidth}
        \centering
        \includegraphics[width=1.0\textwidth]{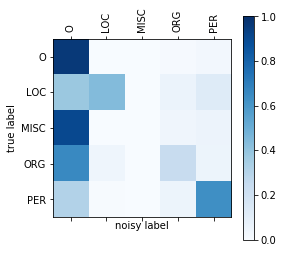}
        \caption{Global Matrix}
        \label{fig:context_global}
    \end{subfigure}
    ~
    \begin{subfigure}[t]{0.19\textwidth}
        \centering
        \includegraphics[width=1.0\textwidth]{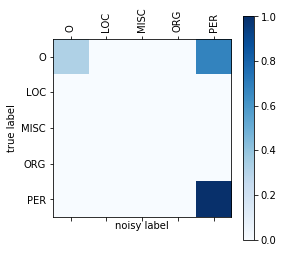}
        \caption{Month Names}
        \label{fig:context_month}
    \end{subfigure}
    ~
    \begin{subfigure}[t]{0.19\textwidth}
        \centering
        \includegraphics[width=1.0\textwidth]{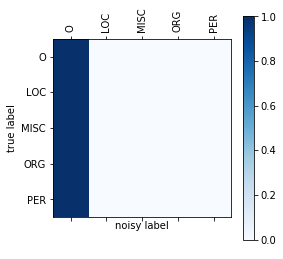}
        \caption{Uppercased Words}
        \label{fig:context_uppercase}
    \end{subfigure}
    ~
    \begin{subfigure}[t]{0.19\textwidth}
        \centering
        \includegraphics[width=1.0\textwidth]{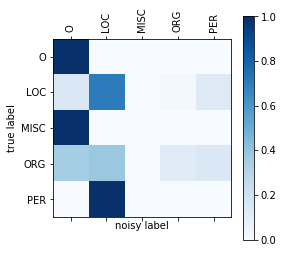}
        \caption{Location Names}
        \label{fig:context_locations}
    \end{subfigure}
    ~
    \begin{subfigure}[t]{0.19\textwidth}
        \centering
        \includegraphics[width=1.0\textwidth]{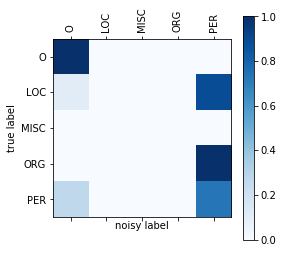}
        \caption{Asian Names}
        \label{fig:context_asian}
    \end{subfigure}
    ~
    \begin{subfigure}[t]{0.19\textwidth}
        \centering
        \includegraphics[width=1.0\textwidth]{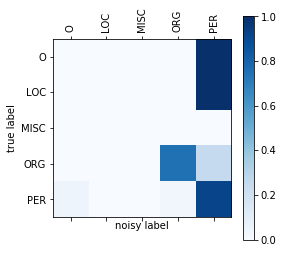}
        \caption{First Names}
        \label{fig:context_firstnames}
    \end{subfigure}
    
    \caption{Confusion matrices used for initialization when training with the English dataset. The global matrix is given as well as five of the feature-dependent matrices obtained when using $k$-Means clustering for 75 clusters. }
    \label{fig:context_matrices}
\end{figure}

\section{Experiments}
We evaluate all models in five low-resource NER settings across different languages. Although the evaluation is performed for NER labeling, the proposed models are not restricted to the task of NER and can potentially be used for other tasks. 

\subsection{Models
\footnote{The code for all models is made available at \url{https://github.com/uds-lsv/noise-matrix-ner}
}}

We follow the BiLSTM architecture from \citet{noise/Hedderich18}. Only the optimizer was changed for all models to NADAM~\cite{nadam/Dozat16} as this helped with convergence problems for increasing cluster numbers. The \textbf{Base} is trained only on clean data while \textbf{Base+Noise} is trained on both the clean and the noisy data without noise handling. \textbf{Global-CM} uses a global confusion matrix for all noisy instances to model the noise as proposed by \citet{noise/Hedderich18} and presented in Section~\ref{sec:global}. The same architecture is used for \textbf{Global-ID-CM}, but the confusion matrix is initialized with the identity matrix (instead of Formula~\ref{eq:global-init}) and only adapted during training.

The cluster-based models we propose in Section~\ref{sec:feature-model} are \textbf{Brown-CM} and \textbf{K-Means-CM}. We experimented with numbers of clusters of 5, 10, 25 and 50. The models that select only the largest groups $G$ are marked as \textbf{*-Freq} and select either 30\% or 50\% of the clusters. The interpolation models have the postfix \textbf{*-IP} with $\lambda \in \{0.3, 0.5, 0.7\}$ . The combination of both is named \textbf{*-Freq-IP}. As for all other hyperparameters, the choice was taken on the development set.

We implemented the \textbf{Cleaning}~\citep{noise/Veit17} and \textbf{Dynamic-CM}~\citep{noise/Luo17} models. Both were not developed for sequence labeling tasks and therefore needed to be adapted. For the Cleaning model, we followed the instructions by \citet{noise/Hedderich18}.
The embedding and prediction components of the \textbf{Dynamic-CM} model were replaced according to our base model. The output of the dense layer was used as input to the dynamic matrix generation. We experimented with and without their proposed trace loss. 

The training for all models was performed with labels in the IO format. The predicted labels for the test data were converted and evaluated in IOB2 with the official CoNLL evaluation script. The IOB2 format would increase matrix size making the confusion matrix estimation more difficult without adding much information in practice. In preliminary experiments, this decreased performance in particular for low-resource settings. 

\begin{table}[b]
    \centering
    \footnotesize
    \begin{tabular}{l|ccccc} \toprule
              & De   & En   & Es   & Et   & Nl   \\ \midrule
    Precision & 23.2 & 39.9 & 51.0 & 59.7 & 32.4 \\
    Recall    & 9.2  & 30.1 & 24.7 & 49.3 & 21.1 \\
    F1        & 13.2 & 34.3 & 33.3 & 54.0 & 25.5 \\ \bottomrule
\end{tabular}
\caption{Results of the automatic labeling method proposed by~\citet{ner/Dembowski2017} on the test data.}
\label{tab:labeling}
\end{table}

\subsection{Data}
The models were tested on the four CoNLL datasets for English, German, Spanish and Dutch~\cite{conll/Sang02, conll/Sang03} using the standard split, and the Estonian data from~\citet{estonian/Tkachenko13} using a 10/10/80 split for dev/test/train sets. 
For each language, the labels of 1\% of the training data (ca. 2100 instances) were used to obtain a low-resource setting. We treat this as the clean data $C$. 
The rest of the (now unlabeled) training data was used for the automatic annotation which we treat as noisily labeled data $N$. We applied the distant supervision method by~\citet{ner/Dembowski2017}, which uses lists and gazetteer information for NER labeling. 
As seen in Table~\ref{tab:labeling}, this method reaches rather high precision but has a poor recall. 
The development set of the original dataset is used for model-epoch and hyperparameter selection, and the results are reported on the complete, clean test set. The words were embedded with the pretrained fastText vectors~\citep{fasttext/Grave18}. The clusters were calculated on the unlabeled version of the full training data. Additionally, the Brown clusters used the language-specific documents from the Europarl corpus~\citep{europarl/Koehn05}.

\section{Experimental Results}
The results of all models are shown in Table~\ref{tab:results}. The newly proposed cluster-based models achieve the best performance across all languages and outperform all other models in particular for Dutch and English. 
The combination of interpolation with the global matrix and the selection of large clusters is almost always beneficial compared to the cluster-based models using only one of the methods.
In general, both clustering methods achieve similar performance in combination with interpolation and selection, except for English, where Brown clustering performs worse than $k$-Means clustering. 
While the Brown clustering was trained on the relatively small Europarl corpus, $k$-Means clustering seems to benefit from the word embeddings trained on documents from the much larger common crawl.

\section{Analysis}
In the majority of cases, a cluster size of 10 or 25 was selected on the development set during the hyperparameter search. Increasing the number of clusters introduces smaller clusters for which it is difficult to estimate the noise matrix, due to the limited training resources. On the other hand, decreasing the number of clusters can generalize too much, resulting in loss of information on the noise distribution. 
For the $\lambda$ parameter, a value of either 0.3 or 0.5 was chosen on the development set giving the group clusters more or equal weight compared to the global confusion matrix. This shows that the feature dependent noise matrices are important and have a positive impact on performance. 

Five confusion matrices for groups and the global matrix in the English data are shown as examples in Figure~\ref{fig:context_matrices}.
One can see that the noise matrix can visibly differ depending on the cluster of the input word. 
Some of these differences can also be directly explained by flaws in the distant supervision method. 
The automatic annotation did not label any locations written in all upper-case letters as locations. 
Therefore, the noise distribution for all upper-cased locations differs from the distribution of other location names (cf. \ref{fig:context_locations} and \ref{fig:context_uppercase}). 
The words April and June are used both as names for a month and as first names in English. This results in a very specific noise distribution with many temporal expressions being annotated as person entities (cf. \ref{fig:context_month}). 
Similar to this, first-person names and also Asian words are likely to be labeled as persons by the automatic annotation method (cf. \ref{fig:context_firstnames} and \ref{fig:context_asian}).

All of these groups show traits that are not displayed in the global matrix, allowing the cluster-based models to outperform the other systems.

\section{Conclusions}
We have shown that the noise models with feature-dependent confusion matrices can be used effectively in practice. These models improve low-resource named entity recognition with noisy labels beyond all other tested baselines. Further, the feature-dependent confusion matrices are task-independent and could be used for other NLP tasks, which is one possible direction of future research.

\section*{Acknowledgments}
The authors would like to thank Heike Adel, Annemarie Friedrich, Jannik Str\"{o}tgen and the anonymous reviewers for their helpful comments. This work has been partially funded by Deutsche Forschungsgemeinschaft (DFG) under grant SFB 1102: Information Density and Linguistic Encoding.

\bibliography{emnlp-ijcnlp-2019}
\bibliographystyle{acl_natbib}

\end{document}